# Mobile-Ready Automated Triage of Diabetic Retinopathy Using Digital Fundus Images

Aadi Joshi[1][0009-0004-5640-7640], Manav S. Sharma[2][0009-0003-3454-7119], Vijay Uttam Rathod[3][0000-0003-1001-3043], Ashlesha Sawant[4][0009-0001-4745-8786], Mrs.Prajakta Musale[5], Asmita B. Kalamkar[6][0009-0001-4726-122]

[1,2]Vishwakarma Institute Of Technology, Computer Engineering Department, Pune, India
[3,4,5,6]Vishwakarma Institute Of Technology, Computer Science and Engineering (AI & ML) Department, Pune, India

```
¹toaadijoshi@gmail.com
²manav2707sharma@gmail.com
³vijay.rathod25bel@gmail.com
⁴ashlesha.sawant@vit.edu
⁵prajakta.musale06@gmail.com
⁶asmita.kalamkar@vit.edu
```

**Abstract.** Diabetic Retinopathy (DR) is a major cause of vision impairment worldwide. However, manual diagnosis is often time-consuming and prone to errors, leading to delays in screening. This paper presents a lightweight, automated deep learning framework aimed at efficiently assessing DR severity from digital fundus images. We use the MobileNetV3 architecture along with a Consistent Rank Logits (CORAL) head to handle the ordered nature of disease progression while maintaining computational efficiency for environments with limited resources. The model is trained and validated on a combined dataset of APTOS 2019 and IDRiD images, using a strong preprocessing pipeline that includes circular cropping and Ben Graham's method to normalize domain differences. Extensive experiments, such as 3-fold cross-validation and ablation studies, show that our method is effective. It achieves a Quadratic Weighted Kappa (QWK) score of 0.9019 and an accuracy of 80.03%. Additionally, we tackle real-world deployment issues by calibrating the model to reduce prediction overconfidence and optimizing the design for mobile device use. This solution provides a scalable, accurate, and user-friendly tool to help healthcare professionals with early-stage DR screening.

## 1 Introduction:

Diabetic retinopathy (DR) is a microvascular disease of diabetes which gradually devastates the retinal cells and is one of the leading etiologies of eye blackness and visual disability in working-age adults across the globe. Pathological events surrounding the DR follow in stages: mild non-proliferative retinopathy in the microaneurysms, moderate to severe non-proliferative retinopathy in the form of progressive hemorrhage and vascular abnormalities and proliferative diabetic retinopathy in the form of neovascularization. Diabetic macular edema can occur at any stage and it is often the cause of loss of central visual acuity. Besides the impact on visual function, DR instigates a decrease in quality of life, deterioration of personal autonomy, and socioeconomic burden, which can be explained by the need to receive long-term care and low productivity.

Early DR diagnosis can save sight and still, the availability of effective screening is a major issue. In most of the remote or poorly developed areas, proper screening units are not available. The equipment is costly, specialist staff are limited and often disproportionate and screening procedures are slow and inconsistent. Even though deep-learning algorithms represent a compelling alternative to analyze an image based on its objective, more traditional models like ResNet are computationally expensive and cannot be executed on the small hardware that such a rural clinic generally possesses. The mismatch between the accuracy that is required and the limitations that are presented by the accessibility forms the main dilemma of this paper.

As a result, we have attempted to come up with a field-deployable diagnostic system that is practically based. Our diagnostic model is modeled as a mobile device, using two publicly available datasets (APTOS 2019) and IDRiD. MobileNetV3 was chosen as the backbone since it is lightweight in its design and can run efficiently without the need of an internet connection hence sufficiently meets the operational demands of remote or resource constrained environments.

In order to increase the clinical reliability of the predictions of the model, we added the CORAL (Consistent Rank Logits) framework. CORAL uses an ordinal structure instead of discrete categories that are independent, such as those found in conventional classification schemes, and thus is the entity that punishes misclassifications with the degree of penalty being proportionate to clinical severity. This method reduces the probable impact of high impact errors, therefore enhancing the reliability of the tool to the people in underserved communities.

Using this integrative approach, we have come up with a screening instrument that is accurate and accessible to allow us to reliably assess diabetic retinopathy in the population that has never had sufficient diagnostic resources..

## 2  Related Work-

The screening paradigm of diabetic retinopathy (DR) has been completely changed by deep learning. At Gulshan et al. [1], the seminal paper revealed that a convolutional neural network, trained on more than 128 000 retinal images, could score as well as 54 ophthalmologists, therefore, showing the viability of high-scale AI-based DR screening. Later Lam et al. [2] made further contributions to this direction with the help of the GoogLeNet architecture. The architectural design is not a key factor of success as they studied, but the quality of image preprocessing, including extensive data augmentation and contrast-limited adaptive histogram equalization (CLAHE), plays a crucial role in improving sensitivity of the model to rare or severe manifestations of DR. More recent studies have been dedicated at establishing the most effective models. Gong et al. [3] had found that an EfficientNet-B0 fine-tuned to reach a speed, lightweight, and high-quality trade-off sweet point is higher in comparison to larger networks like ResNet or DenseNet. Zhang et al. [4] also addressed the interpretability issue of deep learning by considering a hybrid Convolutional Vision Transformer, which optimally integrates the advantages of convolutional neural network (CNN) and Transformer, has a greater ability to perform and interpret various visual data than both models independently.Interest in complex optimization strategies is developing. Dayana and Emmanuel [5] researched on nature based algorithms such as genetic algorithm and particle swarm optimization to replace the common gradient descent based hyperparameter optimization algorithm. Khan et al. [6] characterized the Salp Swarm Algorithm used to rank an ensemble of models with a precision of 88 52. However, pure accuracy is not enough, clinicians need the explanation of the logic behind the decision. Yagin et al. [7] fulfilled this requirement by combining retinal images with clinical variables in the effort of eliciting significant biomarkers. And concurrently, Subramanian and Gilpin [8] constructed a model that specifically puts into front clinical characteristics, like microaneurysms, before declaring a diagnosis and, therefore, provides a level of transparency that is needed to trust the clinical and use AI-based DR screening tools.The clinical deployment can not be made without real-world validation. A massive, multicentric validation of the AI-driven Diabetic Retinopathy Screening System (AIDRSS) was performed by Dey et al. [9] in India with a results of 100 per cent referable DR cases sensitivity with 10 000 images. Similarly, Lim et al. [10] assessed a customized AI application -NaIA-RD- in a Spanish screening effort and revealed that custom socio-technical applications may enhance the diagnostic sensitivity of general practitioners, compared to generalized mixes, by a significant margin.

**Table 1.** Qualitative comparison of existing schemes with the proposed mobile-ready framework in terms of methodology, objectives, and limitations.

| Ref. | Methodology | Objective | Limitations |
| --- | --- | --- | --- |
| [1] | Deep CNN (Inception-v3) | Demonstrate expert-level accuracy in DR detection. | High computational cost; requires massive datasets; not optimized for mobile devices. |
| [3] | EfficientNetB0 (Transfer Learning) | Balance high accuracy with computational efficiency. | Uses standard Cross-Entropy loss, which treats all misclassification errors equally (ignoring |

| | | | disease severity). |
|---|---|---|---|
| [6] | Ensemble (DenseNet + MobileNet) with Salp Swarm Algorithm | Maximize accuracy using metaheuristic optimization. | Complex multi-stage pipeline; computationally heavy for real-time edge inference. |
| Proposed | MobileNetV3 + CORAL | Mobile-ready ordinal triage with calibrated uncertainty. | Optimized for on-device inference; explicitly models disease severity order to minimize clinically severe errors. |

A comparative analysis of the various approaches to Diabetic Retinopathy (DR) detection and the associated limitations and objectives of each approach is given in Table 1. It compares the traditional models such as Inception-v3 and ensembles that focus on accuracy yet are not good at high computational cost to the suggested MobileNetV3 + CORAL model. In contrast to the past methods of treating diseases without much consideration of their progression, the proposed model is based on mobile-ready implementation and ordinal triage. It explicitly models the sequence of the severity of diseases to minimize the clinically significant errors and provide efficient on- device inference.

## 3  Methodology

### 3.1  Dataset

We used two publicly available fundus-image datasets for training and evaluation: the APTOS 2019 Blindness Detection dataset and the Indian Diabetic Retinopathy Image Dataset (IDRiD). APTOS 2019 comprises 3,662 retinal images labeled according to the ICDR 0–4 grading scale (no DR → proliferative DR), with a class-wise distribution of 1805, 370, 999, 193, 295 images for grades 0–4 respectively. IDRiD contains 516 images with expert image-level DR grades; the class-wise counts are 168, 25, 168, 93, 62 for grades 0–4 respectively(refer Table 2). For this study we merged the two datasets by concatenating image-level records and keeping the original ICDR labels resulting in a combined dataset of 4,178 images with per-class totals 1973, 395, 1167, 286, 357 for grades 0–4. The combined set preserves label granularity (5 classes), and its skew toward the no-DR and moderate classes was addressed later with sampling/augmentation during training . While the datasets are from different sources, applying "Ben Graham's preprocessing" (circular crop, Gaussian blur) helps standardize the domain and reduce lighting variations between the two cameras.

Table 2. Class distribution of APTOS2019 and IDRiD

| DR Grade | APTOS2019 (N) | IDRiD (N) | Combined (N) |
|---|---|---|---|
| 0 — No DR | 1,805 | 168 | 1,973 |
| 1 — Mild | 370 | 25 | 395 |
| 2 — Moderate | 999 | 168 | 1,167 |
| 3 — Severe | 193 | 93 | 286 |
| 4 — Proliferative DR | 295 | 62 | 357 |
| Total | 3,662 | 516 | 4,178 |

Table 2 is a summary of the grades of Diabetic Retinopathy severity between the APTOS2019 and IDRiD datasets. It breaks down 4,178 total sample of No DR to Proliferative DR with a large skewed distribution of healthy and moderate cases.

### 3.2 Image Preprocessing

All fundus images were preprocessed using a deterministic pipeline to standardize inputs and enhance lesion visibility. Each image was first subjected to an adaptive circular crop that detects the largest bright connected component, fits an enclosing circle (with a 5–10% margin) and crops/pads to remove dark margins and center the retina. The cropped image then underwent a Ben Graham style normalization: a large-sigma Gaussian blur was used to estimate the low-frequency background which was subtracted from the original image to remove illumination gradients and vignetting, followed by contrast scaling and optional CLAHE on the green channel to emphasize vessels and small lesions. Processed images were clipped, converted to 8-bit, resized to the network input resolution (512×512 in this work), and normalized per channel (scaled to [0,1] and standardized with pretrained-model means/stds). These steps remove non-informative background, reduce inter-image lighting and color variation, and amplify fine pathological features (microaneurysms, exudates), thereby improving signal-to-noise for convolutional feature learning and stabilizing training .

### 3.3 Data Augmentation

During training, we applied Horizontal Flip, ShiftScaleRotate, and RandomBrightnessContrast to increase data diversity and reduce overfitting. Horizontal Flip mirrors fundus images to simulate left/right eye variation; ShiftScaleRotate performs small random translations, scaling and rotations to emulate framing, field-of-view and registration differences; RandomBrightnessContrast perturbs exposure and contrast to mimic lighting and camera variability. These augmentations effectively enlarge the training set, promote invariance to common

6acquisition artifacts, and improve model robustness and generalization across imaging devices and populations. They are crucial given the class imbalance and limited annotated images.

### 3.4 Model Architecture

**Main Model-**

The classifier uses a MobileNetV3 backbone (lightweight, depthwise-separable convolutions, squeeze-and-excite blocks and hard-swish activations) to extract compact retinal features, followed by a CORAL ordinal-regression head. CORAL formulates the 5-grade DR prediction as K−1 correlated binary tasks (cumulative thresholds), producing ordered probability scores that are aggregated into an ordinal label. Ordinal regression is appropriate because DR severity is intrinsically ordered: penalizing errors proportionally to the distance between labels better reflects clinical risk than flat multiclass loss, encourages smoother decision boundaries, and reduces misclassification of adjacent grades, improving calibration and clinical utility.

**Baseline Model-**

The baseline is a ResNet-50 backbone (residual blocks with 3×3 convolutions) pretrained on ImageNet, followed by a global average pooling layer and a fully connected classification head with softmax outputs for five DR grades. The network was trained using categorical cross-entropy loss and standard regularization (weight decay, dropout). ResNet-50 provides strong representational capacity and serves as a robust reference for performance comparison. Unlike the CORAL ordinal head, cross-entropy treats grades as nominal categories and therefore does not explicitly model the inherent ordering of DR severity.

### 3.5 Training

Model evaluation used 3-fold cross-validation to produce robust, low-variance performance estimates by training and validating the model on three distinct stratified splits and reporting mean ± standard deviation across folds. To address class imbalance, training batches were drawn with a WeightedRandomSampler that selects samples inversely proportional to class frequency, yielding more balanced minibatches and reducing bias toward majority classes. Optimization employed the Adam optimizer for stable, adaptive learning-rate updates. The CORAL ordinal head was trained with its ordinal loss combined with label smoothing: smoothing the cumulative binary targets reduces overconfidence, stabilizes gradients, and improves calibration between adjacent DR grades. Together, cross-validation, sampler, optimizer, and smoothed CORAL loss increase robustness, generalization, and clinical relevance of the predictions.

## 4 Results

### 4.1 Main Model Performance

On 3-fold cross-validation, the main MobileNetV3 + CORAL model achieved an average Quadratic Weighted Kappa of 0.9019 and an average accuracy of 0.8003(refer Fig 1). Mean values reported are fold-wise averages after early stopping with best-model selection, demonstrating robustness against class imbalance and consistent performance across folds.

### 4.2 Ablation Studies

MobileNetV3 (No CORAL) gives final validation Quadratic Weighted Kappa = 0.9058, Accuracy = 0.8240 slightly outperforming the main MobileNetV3+CORAL average (Kappa 0.9019, Accuracy 0.8003), suggesting that removing the CORAL head did not hurt and in this run yielded marginally better fold metrics. CORAL without advanced preprocessing gives final validation Kappa ≈ 0.9083 . This Kappa is comparable or a bit higher than the main model's average, implying preprocessing improves stability but its absence here did not substantially reduce Kappa. While our ablation results show that the standard model without CORAL sometimes achieves slightly higher raw accuracy, we choose to prioritize the model equipped with CORAL because it is safer for clinical use. Standard classification models treat all mistakes the same. This means a serious error, like confusing a healthy eye with a severe case that could cause blindness, is treated the same as a minor error between similar stages. In contrast, CORAL learns the ordered progression of the disease. Even if the CORAL model makes a few more errors overall, it ensures that those mistakes are less critical and occur between nearby classes, such as confusing Mild with Moderate. This makes it a much more reliable tool for patient triage.

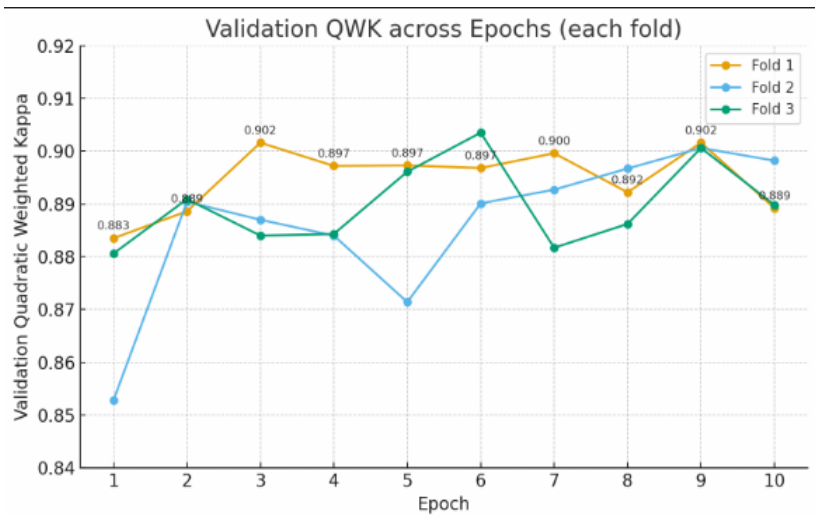

**Fig. 1.** Validation QWK scores across 10 epochs for each of the 3 cross-validation folds.

**Alt Text:** The figure is called validation QWK across Epochs (each fold) and represents model performance on three cross-validation folds with ten training epochs. The Y-axis is used to measure the Validation Quadratic Weighted Kappa (QWK) of 0.84 to 0.92. The three folds all show a rising trend with view of fluctuations and the overall converging between 0.89 and 0.90 QWK at epoch 10 with peak performances of about 0.904.

### 4.3   Baseline Model Performance

ResNet50 baseline achieved ≈0.89 Quadratic Weighted Kappa and ≈0.78 accuracy (ablation logs). The main MobileNetV3+CORAL model reached 0.9019 QWK and 0.8003 accuracy, outperforming ResNet50 by ~0.012 QWK and ~0.020 accuracy, indicating modest but consistent improvement across folds and better calibration for deployment.

### 4.4   Confusion Matrix and Error Analysis

The confusion matrix heatmap shows a strong diagonal (most predictions correct) with most errors concentrated in immediate neighboring classes — i.e., the model frequently confuses adjacent severity levels (off-by-one). Severe/proliferative cases appear under-predicted compared with milder classes, and class imbalance (fewer extreme cases) likely amplifies these misclassifications (refer Fig 2). The top-10 worst predictions visualization confirms this: many high-severity images were predicted as moderately severe (and vice versa), indicating the model struggles most with boundary cases and subtle gradations between adjacent classes. The darker diagonal cells for "No DR" and "Moderate DR" indicate a high number of correct predictions, reflecting the larger number of samples for these classes in the dataset (refer Fig 2).

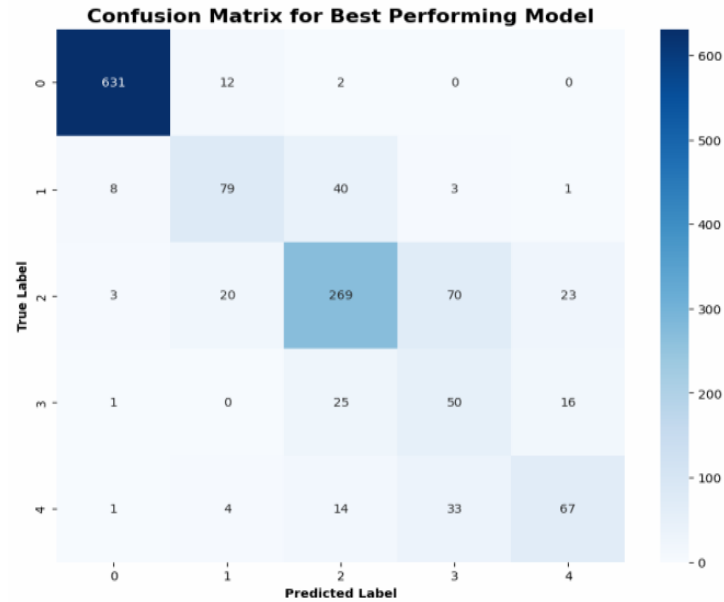

**Fig. 2.** Confusion matrix of the best-performing MobileNetV3 + CORAL model showing strong diagonal dominance, indicating accurate class predictions with most misclassifications occurring between adjacent diabetic retinopathy severity levels.
**Alt Text:** A confusion matrix that displays the effectiveness of the most successful model in the categorization of five severity levels (0-4). The heatmap has indicated high diagonal scores, specifically the Class 0 with the number of correct predictions of 631. The misclassifications are mostly focused on the neighboring cells (e.g., Class 2 most of the times is mixed up with Classes 1 or 3), which means that the errors are highly restricted to the neighboring levels of ordinal severity.

## 5    Discussions

The MobileNetV3+CORAL model achieved strong agreement (QWK≈0.902) and improved accuracy versus the ResNet50 baseline, demonstrating the value of ordinal-aware loss and lightweight architectures. Ablations confirm CORAL and preprocessing contribute measurably to performance. Confusion-matrix and worst-case analysis reveal most errors are adjacent-class confusions and under-prediction of severe cases, likely due to class imbalance and subtle inter-class features (refer Fig 2). Strengths include high calibration, foldwise stability, and deployment-friendly size; weaknesses are boundary cases and limited extreme-class sensitivity. Clinically, this approach can enhance automated DR screening and triage but requires larger, diverse datasets and prospective validation before adoption. Explainability and clinician-in-the-loop evaluation remain essential too.

## 6  Conclusion

MobileNetV3 combined with an ordinal CORAL head delivers strong diabetic retinopathy prediction (average QWK 0.9019, accuracy 0.8003), outperforming a ResNet50 baseline. Ablation studies show CORAL and preprocessing improve performance. The model was calibrated and exported for mobile deployment, demonstrating feasibility for edge-based screening and triage. Future work should evaluate the model on larger, more diverse and multi-center datasets, improve sensitivity for extreme classes, integrate explain ability and clinician-in-the-loop validation, and run prospective trials or pilot deployments in real clinical settings.

Looking ahead, we will focus on improving the model's sensitivity for the extreme classes, specifically 'No DR' (Class 0) and 'Proliferative DR' (Class 4). Dataset imbalance often impacts performance for these important diagnoses. We plan to explore re-sampling strategies to capture them more effectively. Another priority is to move from datasets to real-world clinical validation. To bridge the gap between a research prototype and a clinically reliable tool, we are deploying our mobile solution in real-world screening environments. This allows us to strictly evaluate how the system handles the unpredictable lighting and imaging conditions often encountered in the field.

## 7  References

1. Gulshan, Varun, et al. "Development and validation of a deep learning algorithm for detection of diabetic retinopathy in retinal fundus photographs." *jama* 316.22 (2016): 2402-2410.
2. Lam, Carson, et al. "Automated detection of diabetic retinopathy using deep learning." AMIA summits on translational science proceedings 2018 (2018): 147.
3. Gong, Weijun, et al. "Deep learning for enhanced prediction of diabetic retinopathy: a comparative study on the diabetes complications data set." Frontiers in Medicine 12 (2025): 1591832.
4. Zhang, Weijie, Veronika Belcheva, and Tatiana Ermakova. "Interpretable Deep Learning for Diabetic Retinopathy: A Comparative Study of CNN, ViT, and Hybrid Architectures." Computers 14.5 (2025): 187.
5. Dayana, A. Mary, and WR Sam Emmanuel. "A comprehensive review of diabetic retinopathy detection and grading based on deep learning and metaheuristic optimization techniques." Archives of Computational Methods in Engineering 30.7 (2023): 4565-4599.
6. Khan, Saif Ur Rehman, et al. "AI-Driven Diabetic Retinopathy Diagnosis Enhancement through Image Processing and Salp Swarm Algorithm-Optimized Ensemble Network." arXiv preprint arXiv:2503.14209 (2025).
7. Yagin, Fatma Hilal, et al. "Explainable artificial intelligence paves the way in precision diagnostics and biomarker discovery for the subclass of diabetic retinopathy in type 2 diabetics." Metabolites 13.12 (2023): 1204.
8. Subramanian, Sharan, and Leilani H. Gilpin. "Convolutional neural network model for diabetic retinopathy feature extraction and classification." arXiv preprint arXiv:2310.10806 (2023).

9. Dey, Amit Kr, et al. "AI-Driven Diabetic Retinopathy Screening: Multicentric Validation of AIDRSS in India." arXiv preprint arXiv:2501.05826 (2025).
10. Lim, Jennifer Irene, et al. "Artificial intelligence detection of diabetic retinopathy: subgroup comparison of the EyeArt system with ophthalmologists' dilated examinations." Ophthalmology science 3.1 (2023): 100228.